\documentclass[conference]{IEEEtran}

\usepackage[acronym]{glossaries} 

\newacronym{adlif}{ADLIF}{adaptive leaky integrate and fire}
\newacronym{ann}{ANN}{Artificial Neural Network}
\newacronym{asic}{ASIC}{application specific integrated circuit}
\newacronym{bp}{BP}{backpropagation}
\newacronym{bptt}{BPTT}{backpropagation through time}
\newacronym{dcls}{DCLS}{dilated convolution with learnable spacings}
\newacronym{decolle}{DECOLLE}{Dynamics for Deep Continuous Local Learning}
\newacronym{e-prop}{e-prop}{Eligibility Propagation}
\newacronym{kws}{KWS}{keyword spotting}
\newacronym{lif}{LIF}{leaky integrate and fire}
\newacronym{li}{LI}{leaky integrate}
\newacronym{ostl}{OSTL}{Online Spatio-Temporal Learning}
\newacronym{rad}{RAD}{rectified axonal delay}
\newacronym{rtrl}{RTRL}{Real-Time Recurrent Learning}
\newacronym{sc}{SC}{speech commands}
\newacronym{slayer}{SLAYER}{spike layer error reassignment in time}
\newacronym{snn}{SNN}{Spiking Neural Network}
\newacronym{srnn}{SRNN}{spiking recurrent neural network}
\newacronym{shd}{SHD}{spiking heidelberg digits}
\newacronym{sram}{SRAM}{Static Random Access Memory}
\newacronym{srm}{SRM}{spike response model}
\newacronym{ssc}{SSC}{spiking speech commands}
\newacronym{ttfs}{TTFS}{time-to-first-spike}
\newacronym{swap}{SWaP}{size weight and power}
\newacronym{vad}{VAD}{variable axonal delay}

\newacronym{radlif}{RadLIF}{recurrent adaptive leaky integrate and fire}
\newacronym{cadlif}{c-AdLIF}{constrained-\gls{adlif}}

\usepackage{adjustbox}              
\usepackage{amsmath}
\usepackage{cite}					
\usepackage{cleveref}
\usepackage{comment}  				

\usepackage{enumitem}				
\usepackage{graphicx}
\graphicspath{{./figures/}}		
\usepackage{multirow}				
\usepackage{tabularx} 				
\usepackage{threeparttable}
\usepackage{subfig}
\usepackage{soul} 					
\usepackage{xargs}    				
\usepackage[svgnames]{xcolor}			

\IEEEoverridecommandlockouts
\usepackage{cite}
\usepackage{amsmath,amssymb,amsfonts}
\usepackage{algorithmic}
\usepackage{graphicx}
\usepackage{textcomp}
\usepackage{xcolor}

\usepackage{microtype}
\usepackage{graphicx}
\usepackage{booktabs} 

\usepackage{amsmath}
\usepackage{amssymb}
\usepackage{mathtools}
\usepackage{amsthm}

\usepackage{caption}
\usepackage[utf8]{inputenc}
\usepackage{multirow}

\setacronymstyle{long-short}
\makeglossaries

\usepackage[a4paper, total={184mm,239mm}]{geometry}
\def\BibTeX{{\rm B\kern-.05em{\sc i\kern-.025em b}\kern-.08em
    T\kern-.1667em\lower.7ex\hbox{E}\kern-.125emX}}
    
\begin{document}

\title{Three-Factor Delay Learning Rules Spiking Neural Networks}

\author{\IEEEauthorblockN{Luke Vassallo and Nima Taherinejad}
\IEEEauthorblockA{\textit{Institute for Computer Engineering} \\
\textit{Heidelberg University}\\
Heidelberg, Germany \\
\{luke,nima\}@ziti.uni-heidelberg.de}
}

\maketitle

\begin{abstract}
\Glspl{snn} are dynamical systems that operate on spatiotemporal data, yet their learnable parameters are often limited to synaptic weights, contributing little to temporal pattern recognition. Learnable parameters that delay spike times can improve classification performance in temporal tasks, but existing methods rely on large networks and offline learning, making them unsuitable for real-time operation in resource-constrained environments. In this paper, we introduce synaptic and axonal delays to \gls{lif}-based feedforward and recurrent \glspl{snn}, and propose three-factor learning rules to simultaneously learn delay parameters online. We employ a smooth Gaussian surrogate to approximate spike derivatives exclusively for the eligibility trace calculation, and together with a top-down error signal determine parameter updates. Our experiments show that incorporating delays improves accuracy by up to 20\% over a weights-only baseline, and for networks with similar parameter counts, jointly learning weights and delays yields up to 14\% higher accuracy. On the SHD speech recognition dataset, our method achieves similar accuracy to offline backpropagation-based approaches. Compared to state-of-the-art methods, it reduces model size by 6.6$\times$ and inference latency by 67\%, with only a 2.4\% drop in classification accuracy. Our findings benefit the design of power and area-constrained neuromorphic processors by enabling on-device learning and lowering memory requirements.
\end{abstract}

\glsresetall

\section{Introduction}
\glspl{snn} are continuous time dynamical systems that integrate a weighted sum of action potentials and emit a spike when sufficiently stimulated. Unlike \glspl{ann}, which perform static data transformations without an explicit temporal component, \glspl{snn} inherently operate in the time domain. However, despite this fundamental difference, \glspl{snn} still rely primarily on synaptic weights for learning \cite{bittar_surrogate_2022, yin_accurate_2021}. The absence of dedicated mechanisms for capturing temporal dynamics often leads to lower task performance, as shown by state-of-the-art methods for learning temporal delays \cite{deckers_co-learning_2024, hammouamri2024learning, sun_learnable_2023}. Alternatively, achieving competitive performance often requires significantly larger models with orders of magnitude more parameters \cite{bittar_surrogate_2022}. Recently improved techniques for training \glspl{snn} such as surrogate gradient methods \cite{neftci_surrogate_2019} \cite{cramer_heidelberg_2022}, have allowed learning of time-dependent parameters on a large scale such as synaptic delay learning.

\begin{figure}[bt!]
	\centering    \includegraphics[width=.5\textwidth]{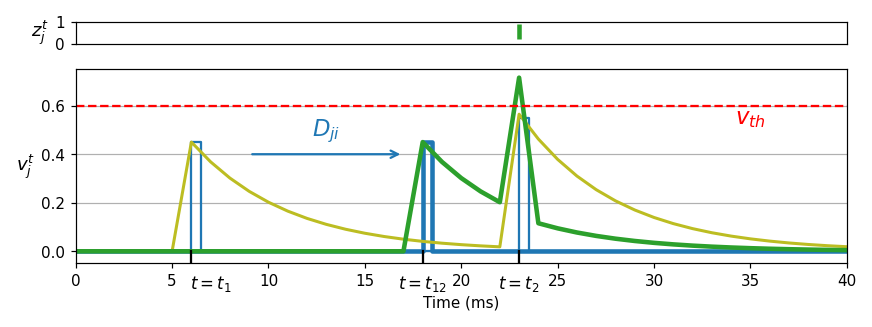} 
	\caption{Illustration of coinciding pre-synaptic spikes leading to post-synaptic. Two distinct pre-synaptic neurons emit spikes at $t=t_1$ and $t=t_2$, respectively. Introducing a delay to the first spike shifts its effect to $t=t_{12}$, where it coincides with the second spike and thereby triggering downstream effects.}
	\label{figure:delay_benefit}
\end{figure}

Learnable delays add a degree of freedom that facilitates temporal pattern detection. For instance, in \Cref{figure:delay_benefit} the green plots shows the effect two incident spikes have on the membrane potential. The first spike at $t_1$ increases the membrane potential in proportion to the strength of the synaptic weight and decays towards zero. A second spike at a later time $t=t_2$ has a similar effect, but since the two do not coincide in time, the membrane potential does not exceed the threshold and thus, the neuron does not fire. However, if $t_1$ is delayed to $t_{12}$ by $D_{ji}$, then the spikes co-incide and, as the bold trace shows, a spikes it emitted. The converse can also be useful, increasing the separation of temporally local spikes can suppress firing.

Our work introduces three-factor delay learning rules for updating synaptic and axonal delays in a gradient-equivalent manner. We explore the application of these rules to \gls{kws} tasks using feedforward \glspl{snn} and \glspl{srnn}. Our results highlight the contribution of each delay type to classification accuracy and demonstrate the advantages of heterogeneous spatiotemporal parameters over homogeneous spatial ones. Our proposed method, along with its systematic evaluation, provides insights for chip designers balancing model versatility and efficiency in \gls{swap}-constrained neuromorphic processors.

\section{Background}
Delays in point neurons can be categorized based on morphology. At the synapse, delays arise from the complex chemical reactions triggered by the arrival of an action potential at the pre-synaptic site, which later initiates the post-synaptic response \cite{sabatini_timing_1996}. If the postsynaptic response causes the neuron to fire, the resulting spike propagates along the axon introducing an axonal delay \cite{debanne_axon_2011}. Before reaching the synapse of a downstream neuron, the signal must also traverse the dendritic arbor, leading to a dendritic delay that depends on dendritic morphology, membrane properties, and the presence of active ion channels \cite{london_dendritic_2005}.

\subsection{Delay learning}
Several approaches exist to embed and learn delays in neural networks. \Gls{slayer} \cite{shrestha_slayer_2018} addresses the temporal credit assignment problem in \glspl{snn}, enabling learning both synaptic weights and delays in multi-layer networks. Sun et~al. \cite{sun_axonal_2022} introduces axonal delays by extending the underlying \gls{srm} with temporal convolutions through \gls{rad} or the neurophysiologically inspired \gls{vad} \cite{sun_learnable_2023}. In both cases, the delay kernel gradient was calculated numerically using the finite difference approximation and demonstrated an accuracy improvement of up to 18.7\% on \gls{kws} datasets \cite{cramer_heidelberg_2022}. Deckers et al. \cite{deckers_co-learning_2024} further adapts \gls{slayer}'s \cite{shrestha_slayer_2018} axonal delay mechanism to individual synapses and when combined with a constrained \gls{adlif} neuron model \cite{bittar_surrogate_2022}, this approach achieves accuracy comparable to temporal convolution methods \cite{hammouamri2024learning}. 

However since in principle the adaptive neuron model and synaptic delays both facilitate temporal pattern recognition, it is unclear how much each feature contributes to classification accuracy. An alternative approach, implemented in PyTorch uses \gls{dcls} \cite{khalfaoui-hassani_dilated_2023, hammouamri2024learning}. It employs a Gaussian kernel parameterized by weight and delay parameters, that is convolved with the incident spike train, to generate the spatiotemporal response of a synapse.

By leveraging automatic differentiation, it achieves state-of-the-art accuracy even on challenging \gls{kws} datasets. Recognizing the benefits of this method in sparse settings, Mészáros et al. \cite{meszaros_learning_2024} introduces dynamic pruning to enforce sparsity. 

Event propagation \cite{wunderlich_event-based_2021} provides an alternative training algorithm that uses spike times to compute exact gradients without surrogates. Applied to keyword spotting and related tasks, event propagation has achieved state-of-the-art performance \cite{meszaros_efficient_2025}, with subsequent experiments highlighting the importance of loss shaping \cite{nowotny_loss_2025}.

\subsection{Online learning}
Backpropagation is update-locked \cite{pmlr-v70-jaderberg17a}, preventing parameter updates until the full input sequence is processed. For sequence data, this results in linearly increasing memory demands, which restricts real-time and on-chip implementation. Online learning releases this constraint and traces its origins to \gls{rtrl} \cite{williams_learning_1989}. By accumulating the gradients of the hidden states, \gls{rtrl} eliminated the need to unroll the network over some time window $T$. However, its cubic memory complexity ($\mathcal{O}(n^3)$) is prohibitive compared to \gls{bptt} ($\mathcal{O}(nT)$). Subsequent research has focused on approximating \gls{rtrl} to reduce computational costs \cite{pmlr-v97-benzing19a, tallec_unbiased_2018} or developing biologically plausible three-factor learning rules \cite{gerstner_eligibility_2018} that aim to retain \gls{bptt}’s performance guarantees.

Three-factor learning rules mitigate the biological implausibility of \gls{bptt}-like algorithms \cite{rumelhart_learning_1986, wunderlich_event-based_2021} by addressing the weight transport \cite{QianLi_2016} and update locking \cite{pmlr-v70-jaderberg17a} problems. They rely on eligibility traces, composed of pre- and post-synaptic factors that capture temporal activity. However, unlike two-factor learning rules \cite{bi_synaptic_1998}, synaptic changes occur only when these traces are modulated by a third factor, for instance, a top-down learning signal spatially propagating task-relevant error information (see \cite{marschall2020} for a comprehensive survey). Superspike \cite{zenke_superspike_2018} constructs an eligibility trace by combining a low-pass filtered trace of pre-synaptic activity with a nonlinear function of post-synaptic activity, while the error signal serves as the third factor. By contrast, \gls{e-prop} \cite{bellec_2020} reformulates \gls{bptt} as the three-factor learning rule framework, demonstrating equivalence for feedforward networks and approximating it for recurrent architectures. \Gls{decolle} \cite{kaiser_synaptic_2020} computes local synthetic gradients using per-layer readout functions. \cite{bohnstingl_2022} decouples spatial gradients (top-down learning signals within the same timestep) and temporal gradients (eligibility traces spanning timesteps). It is equivalent to \gls{bptt} for shallow \glspl{srnn} and approximates its performance for deeper architectures.

\section{Proposed Methods}
We present the neural dynamics for point neurons, network architecture and three-factor delay learning rules using \gls{e-prop} \cite{bellec_2020} as a basis for online learning.

\subsection{Neuron dynamics}
The \gls{lif} model stands out for its computational efficiency and biological plausibility, and has become the defacto standard model in neuromorphic engineering \cite{abbott_lapicques_1999, gerstner_neuronal_2014}. For a point neuron $j$ the \gls{lif} model integrates the weighted sum of input spikes into a one-dimensional hidden variable $v_j^t$, representing the membrane potential. When the membrane potential exceeds a pre-defined threshold, $v_{thr}$ the neuron generates an action potential or spike, $z_j^t \in \{0,1\}$, which propagates along its axon. 

The model consists of two components: a first-order linear differential equation describing the evolution of the membrane potential and a spike generation mechanism. The leaky integration of synaptic currents into the membrane potential $v_j(t)$ is given by 
\begin{equation}
	\label{eqn:lif-ode}
	\tau_m \frac{dv_j}{dt} = -(v_j(t) - v_{reset}) + RI_j(t)
\end{equation}
where $\tau_m$ is the membrane time constant, $ v_{reset}$ is the membrane potential at rest, $R$ is the input resistance, and $I_i(t)$ is the input synaptic current at time, $t$. The synaptic current, described by \Cref{eqn:synaptic_current} is stateless and represented as a weighted sum between the synaptic weight, $W^{in}_{ji}$ and the delayed pre-synaptic input $x_i( t - D^{in}_{ji} )$. The synaptic $D^{in}_{ji} \in \mathbb{Z}$, or $D^{in}_{i} \in \mathbb{Z}$ axonal delay is described in number of timesteps, bounded by $D_{max}$.

\begin{equation}
	\label{eqn:synaptic_current}
	I_j(t) = \sum_i W^{in}_{ji} x_i( t - D^{in}_{ji} )
\end{equation}

\Cref{eqn:lif-ode} is solved analytically under the assumptions $v_{\text{reset}} = 0$ and $R = (1 - e^{-\frac{\Delta t}{\tau_m}})^{-1}$. The solution is discretized with a timestep $\Delta t$, yielding the discrete timestepped neuron dynamics in \Cref{eqn:li}:

\begin{equation}
	\label{eqn:li} 
	v_j^t = \alpha v_j^{t-1} + \sum_j W^{\text{in}}_{ji} x^{t-D^{in}_{ji}}_i
\end{equation}

where $\alpha = e^{-\frac{\Delta t}{\tau_m}}$ is the decay factor determined by the membrane time constant $\tau_m$ and the timestep $\Delta t$.

The spike reset mechanism extends the membrane potential dynamics with a thresholding operation, mathematically represented using the Heaviside function:

\begin{equation}
	\label{eqn:spike_generation}
	z_j^{t+1} = H(v_j^t > v_{thr})
\end{equation}

where $z_j^{t+1}$ is a binary variable indicating the presence of a spike at time $t+1$. Following the spike event, the membrane potential is explicitly reset by deducting $v_{thr}$. The full dynamics for an \gls{snn} are thus given by:

\begin{equation}
	\label{eqn:lif-dly}
	v^{t+1}_j = \alpha v^t_j + W^{in}_{ji} x^{t-D^{in}_{ji}}_i - z^t_j v_{th}
\end{equation}

The neuron dynamics can be extended with recurrent connections yielding an \gls{srnn} with optional delays.

\begin{equation}
	\label{eqn:rlif-dly}
	v^{t+1}_j = \alpha v^t_j + W^{rec}_{ji} z^{t-D^{rec}_{ji}-1}_j + W^{in}_{ji} x^{t-D^{in}_{ji}}_i - z^t_j v_{th}
\end{equation}

\subsubsection{Surrogate gradient}

The spiking mechanism described by \Cref{eqn:spike_generation} is discontinuous and therefore non-differentiable. It blocks the flow of gradient information and is effectively eliminated by means of a surrogate function used only during the calculation of parameter updates. We adopt the piecewise linear function in \Cref{eqn:surrogate_gradient} with $\gamma_{pd}=0.3$.

\begin{equation}
	\label{eqn:surrogate_gradient}
	\frac{dz}{dv_j^t} \approx \frac{\gamma_{pd}}{v_{th}}  \max \left(0,1- \left| \frac{v_{j}-v_{th}}{v_{th}} \right| \right)
\end{equation}

\subsection{Neural network architecture and loss function}
\label{section:arch_and_loss}
The neural network topology consists of a set of virtual neuron, followed a hidden layer without optional recurrent connections, and an output readout layer. Virtual input neurons apply the spiking input signal with an optional axonal delay. The single hidden layer neurons characterized by weights and optional synaptic delays, calculates the neural dynamics as described in \Cref{eqn:lif-dly} for \glspl{snn} and \Cref{eqn:rlif-dly} for \glspl{srnn}.

The \gls{lif} neuron output connects to a readout layer comprised of a set of \gls{li} neurons whose dynamics are also described by \Cref{eqn:li}, however, they omit the spike reset mechanism. The leakage for the readout is defined with by $\kappa = e^{-\frac{\Delta t}{\tau_o}}$, where $\tau_o$ is the membrane time constant.

A large membrane time constant ($\tau_{out} \geq 1s$) is assigned to the \gls{li} neurons in the readout layer, allowing it to practically hold the state indefinitely.

For k-class classification, we assume $k$ categories represented as a K-dimensional one-hot encoded vector. A softmax function is applied to the output from the readout layer, and the cross-entropy loss is computed  $E = - \sum_{k}\sum_{t} \pi_k^t log(\hat{\pi}_k^t)$, where $\pi_k^t$ represents the ground-truth classification label for class $k$ and $\hat{\pi}_k^t$ denotes the corresponding probability predicted by the network.

\subsection{Online learnable delays} 
\acrfull{e-prop} \cite{bellec_2020} is an online learning algorithm that closely approximates \gls{bptt} while describing synaptic weight updates as three-factor learning rules. The formulation supports delay-parametrized spike trains but does not permit the delays to be learnable.  This work proposes the three-factor learning rules in Equation \ref{eqn:e-prop} using the same refactorisation for calculating error gradients with respect to network parameters describing temporal delays. $\frac{dE}{dD_{ji}}$ represents the error gradients with respect to synaptic delays where, $D_{ji} \in \{D^{in}_{ji}, D^{rec}_{ji}\}$. Similarly, $\frac{dE}{dD_{i}}$ corresponds to the gradient with respect to axonal delays, where $D_i \in \{D^{in}_i, D^{rec}_j\}$.

\begin{align}
	\label{eqn:e-prop}
	\frac{dE}{dD_{ji}} &= \sum_t \frac{dE}{dz_j^t} \cdot \left[ \frac{dz_j^t}{dD_{ji}} \right]_{\text{local}},
\end{align}

The top down learning signal $L_j^t = \frac{dE}{dz_j^t}$ is derived from the cross-entropy loss function applied to the readout layer outputs defined in \Cref{section:arch_and_loss}.

The eligibility trace $e^t_{ji} = \frac{\partial z_j^t}{\partial v_j^{t'}}$ incorporates information about the previous spiking activity and can be recursively expressed, permitting real-time implementation as shown in \Cref{eqn:eligibility-trace}. Here, $\frac{\partial z_j^t}{\partial v_j^{t'}}$, is the surrogate gradient of the neuron's output with respect to the hidden state calculated with \Cref{eqn:surrogate_gradient}, $\frac{\partial v_j^t}{\partial v_j^{t-1}}=\alpha$ is computed by differentiating \Cref{eqn:lif-dly} with respect to $v_j^t$, and $\epsilon_{ji}^{t-1}$ is the eligibility vector. 

\begin{equation}
	\label{eqn:eligibility-trace}	
	e_{ji}^t = \frac{\partial z_j^t}{\partial v_j^{t'}} \left( \frac{\partial v_j^t}{\partial v_j^{t-1}} \cdot \epsilon_{ji}^{t-1} + \frac{\partial v_j^t}{\partial D_{ji}} \right)
\end{equation}

Calculating the eligibility trace entails computing the derivative of the hidden state described by \Cref{eqn:rlif-dly,eqn:rlif-dly} with respect to the delay parameters, $\frac{d v_j^t}{dD_{ji}}$. However since $D_{ji}$ is associated with spike trains (input $x^{t-D^{in}_{ji}}_i$ or recurrent $z^{t-D^{rec}_{ji}}_j$) the derivative does not exist. We overcome this problem by representing the spike with a continuous function, \cite{goltz_delgrad_2024, hammouamri2024learning, wang_2019}, particularly the Gaussian kernel in \Cref{eqn:dirac_gauss}.

\subsection{Spike train kernel}
A Gaussian kernel was chosen because preliminary experiments showed that causal kernels, such as the exponential, did not support effective learning, whereas both anti-causal and symmetric kernels did. Between the Gaussian and the double exponential kernels, the Gaussian consistently performed better, likely due to its smoother profile. The non-causal nature of the surrogate spike derivative is undesirable for real-time applications and introduces an implementation cost requiring a ring buffer (see Section \Cref{section:discussion}).

\Cref{eqn:dirac_delay} describes a spike train parametrized with synaptic delay $D_{ji}$ as a sum of $k$ dirac delta functions delayed in time by $t_k + D_{ji}$, 

\begin{equation}
	\label{eqn:dirac_delay}
	x_i^{t-D_{ji}} = \sum_{k}\delta(t-t_k-D_{ji})
\end{equation}

where $\delta$ is the Dirac delta function, $t_k$ are the spike times, and $D_{ji}$ is the synaptic delay between the pre-synaptic neuron $j$ and the post-synaptic neuron $i$. The Dirac delta function is approximated with a Gaussian and parameterized by the synaptic delay, $D_{ji}$ ($D_i$ for axonal delay), where $D_{ji} = -\frac{D_{max}-1}{2}$ represents no delay and $D_{ji} = \frac{D_{max}-1}{2}$ represents the maximum delay, similar to \cite{hammouamri2024learning}. \Cref{eqn:dirac_gauss} describes the delay parametrized Gaussian kernel. During parameter updates, $D_{ji}$ is clamped within $\pm \frac{D_{max}-1}{2}$. 

\begin{equation}
	\label{eqn:dirac_gauss}
	x^{t-D_{ji}}_i \approx \frac{1}{ \sqrt{2\pi} \sigma} e^{-\frac{(t-t_k-D_{ji})^2}{2\sigma^2}}
\end{equation}

The derivative of \Cref{eqn:dirac_gauss} with respect to the delay parameters can now be calculated, yielding:
\begin{equation}
	\frac{d x^{t-D_{ji}}_i}{dt} = - \frac{t-t_k-D_{ji}}{ \sqrt{2\pi} \sigma^3}e^{-\frac{(t-t_k-D_{ji})^2}{2\sigma^2}}    
\end{equation}

\section{Experiments}
Using \gls{kws} tasks \cite{cramer_heidelberg_2022}, we systematically evaluate the proposed three-factor delay learning rules against offline methods \cite{hammouamri2024learning} using dense and sparse \glspl{snn}. The experiments extend to \glspl{srnn}, with ablation tests to identify where delays impact task performance. Hammouamri et. al's \gls{dcls}-based method \cite{hammouamri2024learning} was selected as a baseline because it uses a \gls{lif} model and achieves state-of-the-art performance using temporal convolutions together with \gls{bptt}. Synaptic delays are introduced at input synapses for \glspl{snn} and additionally at recurrent synapses for \glspl{srnn}. Axonal delays are applied to virtual input neurons for \glspl{snn} and to both input and recurrent neurons for \glspl{srnn}.

\subsection{Dataset} 
We use the \gls{shd} and \gls{ssc} datasets to evaluate our implementation of online delay learning. Both datasets consist of spoken sequences in spiking format, generated by preprocessing the original audio using an artificial cochlear model, which converts the auditory signals into spikes \cite{cramer_heidelberg_2022}.
The \gls{shd} dataset consists of 10,000 professionally recorded samples from 12 participants and contains spoken sequences of the digits zero through nine in both English and German. \gls{ssc} dataset is derived from the \gls{sc} dataset \cite{warden2018} and contains 100,000 samples of 35 commonly spoken keywords collected from a diverse group of participants under varying recording conditions.

The datasets consist of 700 spiking neurons with a temporal resolution of 1 ms. To reduce computational time and GPU memory requirements, the spatial dimension is binned by a factor of six, resulting in $700/6=116$ input neurons. Additionally, the temporal dimension is sub-sampled at a rate of 10ms yielding sequences of approximate 100 samples in length.

\subsection{Experimental Setup}
\label{section:experimental_setup}
We evaluate online delay learning in \glspl{snn} and \glspl{srnn} across three configurations: a large fully connected network with 128 hidden neurons, an 80\% sparse variant, and a small fully connected network with 16 hidden units. Sparsity is implemented with fixed random binary maps. Two weights-only configurations, featuring 32 and 64 hidden units, are included to compare performance based on parameter counts. \glspl{snn} experiments are complemented by evaluating the same configurations using \gls{bptt}-based  \gls{dcls}, and for this, we extend the state-of-the-art \cite{hammouamri2024learning} with axonal delays. The online learning experiments are repeated using \glspl{srnn} with both input and recurrent delays (\gls{dcls} experiments are omitted because offline implementation is unsupported for \glspl{srnn}). Evaluations on the \gls{ssc} dataset follow an identical strategy and are limited to 256-neuron fully connected configurations. 

Learnable delays tend to offer a smaller contribution compared to weights \cite{meszaros_learning_2024}. To assess the efficacy of delay learning, we perform control experiments. For \glspl{snn}, we evaluate delay learning as a function of sparsity under fixed and co-learned conditions. For \glspl{srnn}, we fix sparsity and analyze the effect of placing delays at the input, recurrent layer, or both.

The learning rates for weights and delays are set to $10^{-4}$ and $10^{-2}$, respectively. Training is conducted with a batch size of 16 samples over 60 epochs for both the \gls{shd} dataset and \gls{ssc} datasets. Synaptic and axonal delays are limited to 25 timesteps thereby introducing a maximum latency of 250ms. The inference pass uses binary spike trains and the Gaussian kernel is only invoked during parameter updates. Offline \gls{dcls}-based experiments retain all original hyperparameters \cite{hammouamri2024learning}. We train all networks on the training dataset and evaluate top-1 classification accuracy on the test set. Each experiment is repeated five times, and test accuracy is reported with a 95\% confidence interval, assuming a t-distribution. The implementation is performed in PyTorch 2.1 and runs on a Debian 12 system with a Xeon 6526Y CPU, 1 TB of memory, and a 20 GB RTX A4500 Ampere GPU.

\section{Results}
\label{section:results}
\begin{table*}[ht]
	\caption{Test Classification accuracy for \gls{snn} and \gls{srnn} models trained with online and offline (BP) methods.}
	\label{table:our-classification-accuracy}
	\renewcommand{\arraystretch}{1.15} 
	\resizebox{\textwidth}{!}{
		\begin{tabular}{c|c|c|cc|ccc|ccc}
			\hline
			\multirow{2}{*}{ID} & \multirow{2}{*}{Dataset} & \multirow{2}{*}{Configruation} & \multicolumn{2}{c|}{Weights only} & \begin{tabular}[c]{@{}c@{}}Fixed synaptic delays,\\ learnable weights\end{tabular} & \begin{tabular}[c]{@{}c@{}}Learnable weights,\\ and synaptic delays\end{tabular} &            & \begin{tabular}[c]{@{}c@{}}Fixed axonal delays,\\ learnable weights\end{tabular} & \begin{tabular}[c]{@{}c@{}}Learnable weights,\\ and axonal delays\end{tabular} &            \\ \cline{4-11} 
			&                         &                                 & Accuracy (\%)      & Params (k)   & Accuracy (\%)                                                                      & Accuracy (\%)                                                                    & Params (k) & Accuracy (\%)                                                                    & Accuracy (\%)                                                                  & Params (k) \\ \hline
			1 & \multirow{11}{*}{SHD}    & FC$^\dagger$-128 \gls{snn} (BP$^\mathsection$) & 63.70\% ± 1.6\%    & 17.4         & 90.19\% ± 0.27\%                                                                   & 92.79\% ± 1.1\%                                                                  & 32.3       & 74.20\% ± 1.07\%                                                                & 92.02\% ± 1.63\%                                                               & 17.5       \\
			2 &                         & SP$^\ddagger$-128 \gls{snn} (BP)              & 59.72\% ± 3.9\%    & 3.5          & 84.27\% ± 3.32\%                                                                   & 88.42\% ± 2.7\%                                                                  & 6.5        & 48.36\% ± 3.66\%                                                                 & 86.84\% ± 2.84\%                                                               & 3.5        \\
			3 &                         & FC-16 \gls{snn} (BP)                          & 50.33\% ± 6.8\%    & 2.2          & 76.07\% ± 3.79\%                                                                   & 82.79\% ± 1.0\%                                                                  & 4.0        & 56.57\% ± 2.24\%                                                                & 78.96\% ± 6.31\%                                                               & 2.3        \\ \cline{2-11} 
			4 &                         & FC-128 \gls{snn}                                    & 79.46\% ± 0.2\%    & 17.4         & 92.42\% ± 0.22\%                                                                   & 92.64\% ± 0.4\%                                                                  & 32.3       & 91.90\% ± 0.4\%                                                                  & 92.01\% ± 0.3\%                                                                & 17.5       \\
			5 &                         & SP-128 \gls{snn}                                    & 71.23\% ± 1.4\%    & 3.5          & 86.26\% ± 0.79\%                                                                   & 89.22\% ± 0.8\%                                                                  & 6.5        & 86.36\% ± 0.6\%                                                                  & 88.06\% ± 2.2\%                                                                & 3.5        \\
			6 &                         & FC-16 \gls{snn}                                     & 69.09\% ± 1.4\%    & 2.2          & 83.87\% ± 2.47\%                                                                   & 85.58\% ± 4.3\%                                                                  & 4.0        & 84.93\% ± 1.4\%                                                                  & 84.92\% ± 0.7\%                                                                & 2.3        \\
			7 &                         & FC-32 \gls{snn}                                     & 73.88\% ± 1.52\%   & 4.35         & -                                                                                  & -                                                                                & -          & -                                                                                & -                                                                              & -          \\
			8 &                         & FC-64 \gls{snn}                                     & 77.69\% ± 2.14\%   & 8.75         & -                                                                                  & -                                                                                & -          & -                                                                                & -                                                                              & -          \\ \cline{2-11} 
			9 &                         & FC-128 \gls{srnn}                                   & 85.77\% ± 0.94\%   & 33.8         & 91.73\% ± 0.81\%                                                                   & 92.36\% ± 0.3\%                                                                  & 65.0       & 91.66\% ± 0.6\%                                                                  & 91.64\% ± 1.6\%                                                                & 34.0       \\
			10 &                        & SP-128 \gls{srnn}                                   & 83.52\% ± 2.38\%   & 6.76         & 90.18\% ± 0.20\%                                                                   & 91.40\% ± 0.71\%                                                                 & 13.00      & 89.27\% ± 1.40\%                                                                 & 90.42\% ± 0.39\%                                                               & 6.81       \\
			11 &                        & FC-16 \gls{srnn}                                    & 73.29\% ± 1.02\%   & 2.43         & 83.92\% ± 5.80\%                                                                   & 85.81\% ± 2.88\%                                                                 & 4.54       & 84.09\% ± 4.32\%                                                                 & 83.49\% ± 6.58\%                                                               & 2.56       \\ \hline
			12 & \multirow{3}{*}{SSC}     & FC-256 \gls{snn} (BP)                         & 43.16\% ± 0.16\%   & 34.82        & 69.67\% ± 0.10\%                                                                   & 72.87\% ± 0.03\%                                                                 & 64.51      & 43.12\% ± 2.00\%                                                                 & 67.62\% ± 0.30\%                                                               & 34.93      \\
			13 &                        & FC-256 \gls{snn}                                    & 52.24\% ± 0.05\%   & 34.82        & 70.93\% ± 0.24\%                                                                   & 71.90\% ± 0.06\%                                                                 & 64.51      & 69.35\% ± 0.48\%                                                                 & 70.18\% ± 0.04\%                                                               & 34.93      \\
			14 &                        & FC-256 \gls{srnn}                                   & 68.86\% ± 0.05\%   & 100.35       & 73.52\% ± 0.01\%                                                                   & 74.66\% ± 0.09\%                                                                 & 195.58     & 72.37\% ± 0.14\%                                                                 & 73.18\% ± 0.07\%                                                               & 100.72     \\ \hline
		\end{tabular}
	}
	{\footnotesize $^\dagger$ FC = Fully Connected, $^\ddagger$ SP = Sparse (80\%), $^\mathsection$ BP = Backpropagation}    
\end{table*}

Table \ref{table:our-classification-accuracy} presents the top-1 test classification accuracy obtained by \gls{snn} and \gls{srnn} configurations described in \Cref{section:experimental_setup} on the \gls{shd} and \gls{ssc} datasets. Comparing weights-only \glspl{snn} with online learning to the \gls{dcls}-based offline implementation reveals a 15\% improvement in our favor. This large gap was unexpected, but it can likely be explained by our precise analytical \gls{lif} dynamics, rather than the direct Euler integration used in the SpikingJelly-based \cite{Fang2023_spikingjelly} baseline. Recent studies show that this type of discretization can pose challenges in stability and parameterization \cite{baronig_advancing_2025}. For configurations with synaptic and axonal delays, our implementation achieves test accuracies for fully connected models that closely match backpropagation within 0.15\%. This result falls within statistical bounds, demonstrating that our approach is equivalent to offline methods under the same model configurations. In sparse and small network configurations, the inclusion of delays improves performance, surpassing \gls{bptt} by 1\% and 4.3\% respectively.

Temporal delays bring a significant improvement in test classification accuracy. In fully connected networks, accuracy increases by 13.2\% with the addition of synaptic delays and by a slightly lower 12.6\% with axonal delays. The gap widens in sparse and small models, with accuracy gains reaching 18\% and 16.8\% for synaptic and axonal delays, respectively. Under sparse \gls{snn} conditions, learnable delays contribute 3\% and 2.3\% in classification accuracy for synaptic and axonal delays, respectively, compared to fixed random delays. For small networks, the amount is smaller for synaptic delays and negligible for axonal delays.

Simultaneous weight and delay learning improves the model's accuracy-to-parameter ratio. Sparse models with synaptic delays (6.5k parameters) achieve 11.5\% higher classification accuracy than 64-wide fully connected model with 8.75k parameters. Sparse models with axonal delays (3.5k parameters) provide an even greater improvement of 14.2\% over a 32-wide fully connected model with 4.35k parameters. For smaller models, synaptic delays contribute an 11.7\% accuracy increase, whereas axonal delays offer a higher improvement of 15.84\%. Axonal delays bring the highest increase in test classification accuracy per added model parameter. However, the finer granularity provided by synaptic delays consistently results in higher accuracy, particularly in the presence of sparsity.

Three factor learning rules also facilitates training \gls{srnn}. Concerning the \gls{shd} dataset, weights-only \glspl{srnn} establish a baseline at 85.72\% in fully connected configurations, 6.3\% higher than \glspl{snn}. When adding input and recurrent delays, we observe comparable performance with \glspl{snn}, albeit lagging behind by approximately 0.5\% while showing increased robustness in sparse configurations. The classification accuracy on the \gls{shd} dataset seems to saturate at approximately 92.5\% for our setup, despite \glspl{srnn} doubling the number of parameters and adding recurrent connections. 

\begin{figure*}[!ht]
	\centering
	\subfloat[]{%
		\includegraphics[width=0.24\textwidth]{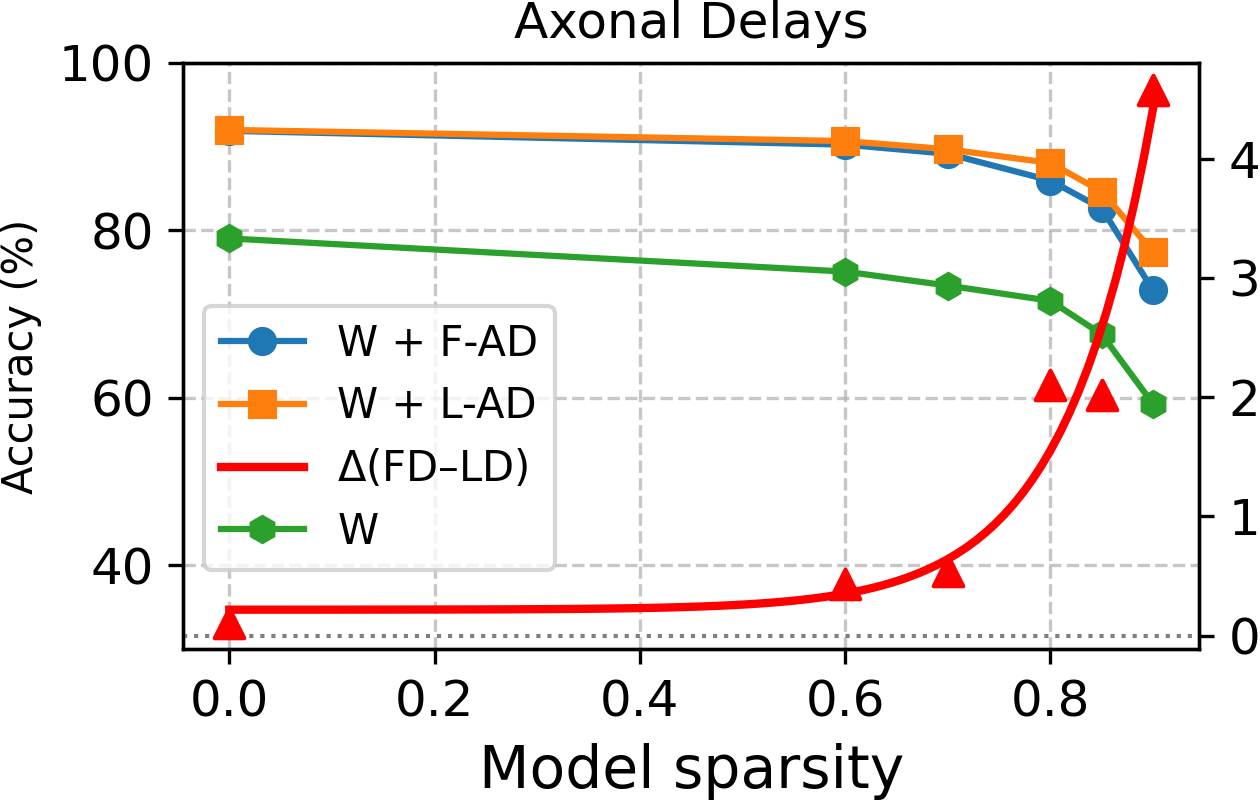}%
		\label{fig:sweep_axonal}}
	\hfill
	\subfloat[]{%
		\includegraphics[width=0.24\textwidth]{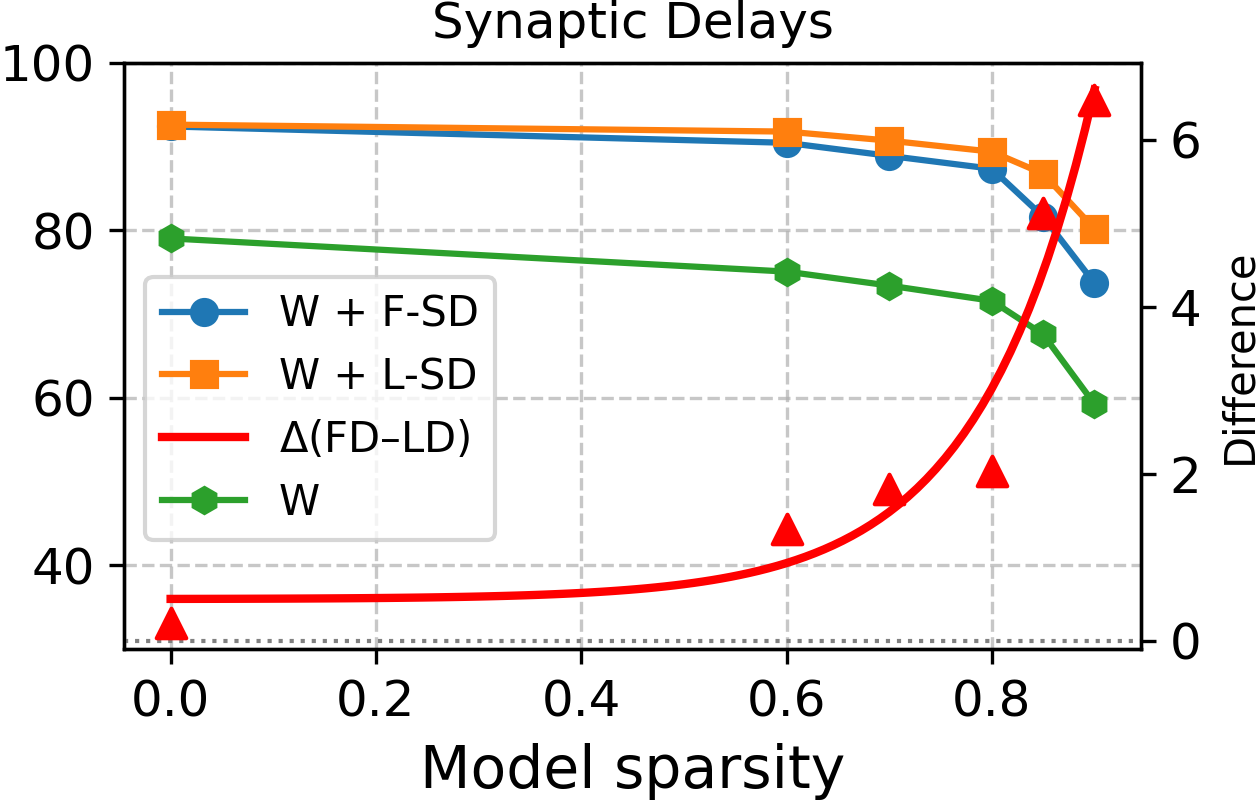}%
		\label{fig:sweep_synaptic}}
	\hfill
	\subfloat[]{%
		\includegraphics[width=0.24\textwidth]{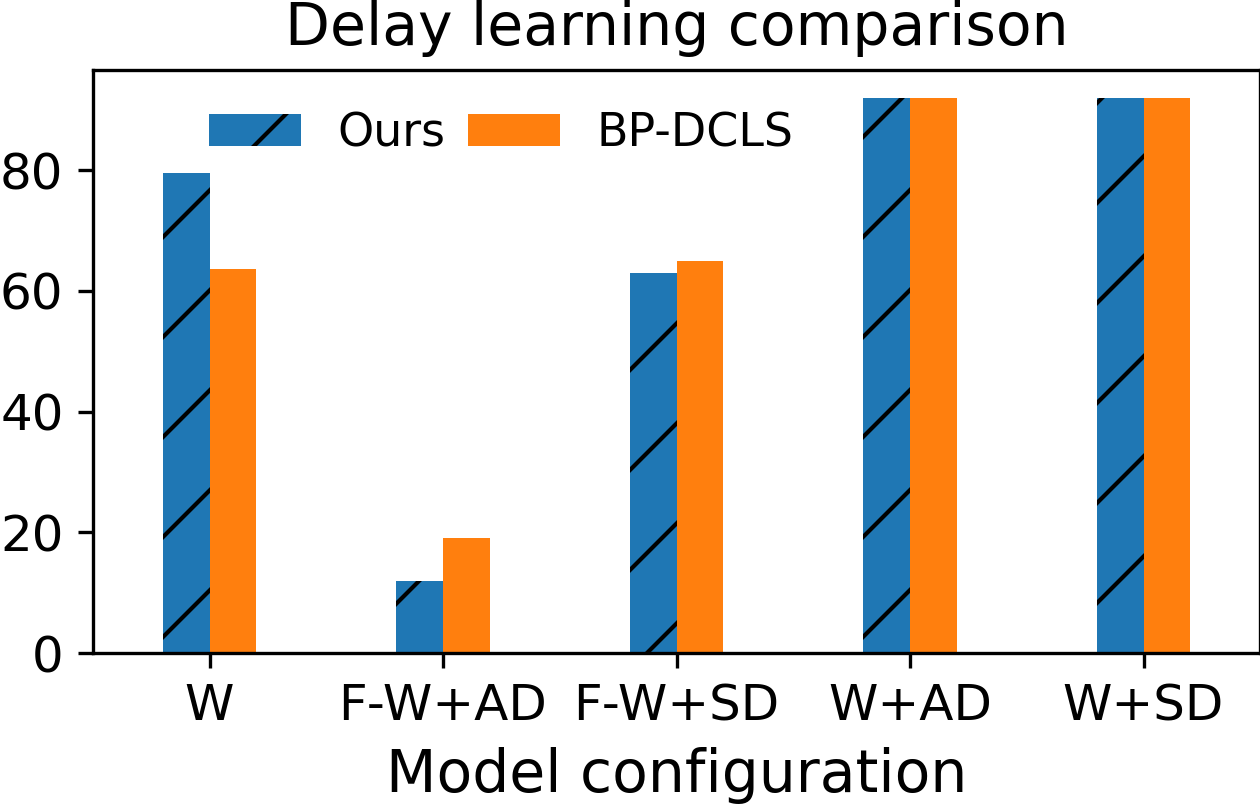}%
		\label{fig:delay_learning}}
	\hfill
	\subfloat[]{%
		\includegraphics[width=0.24\textwidth]{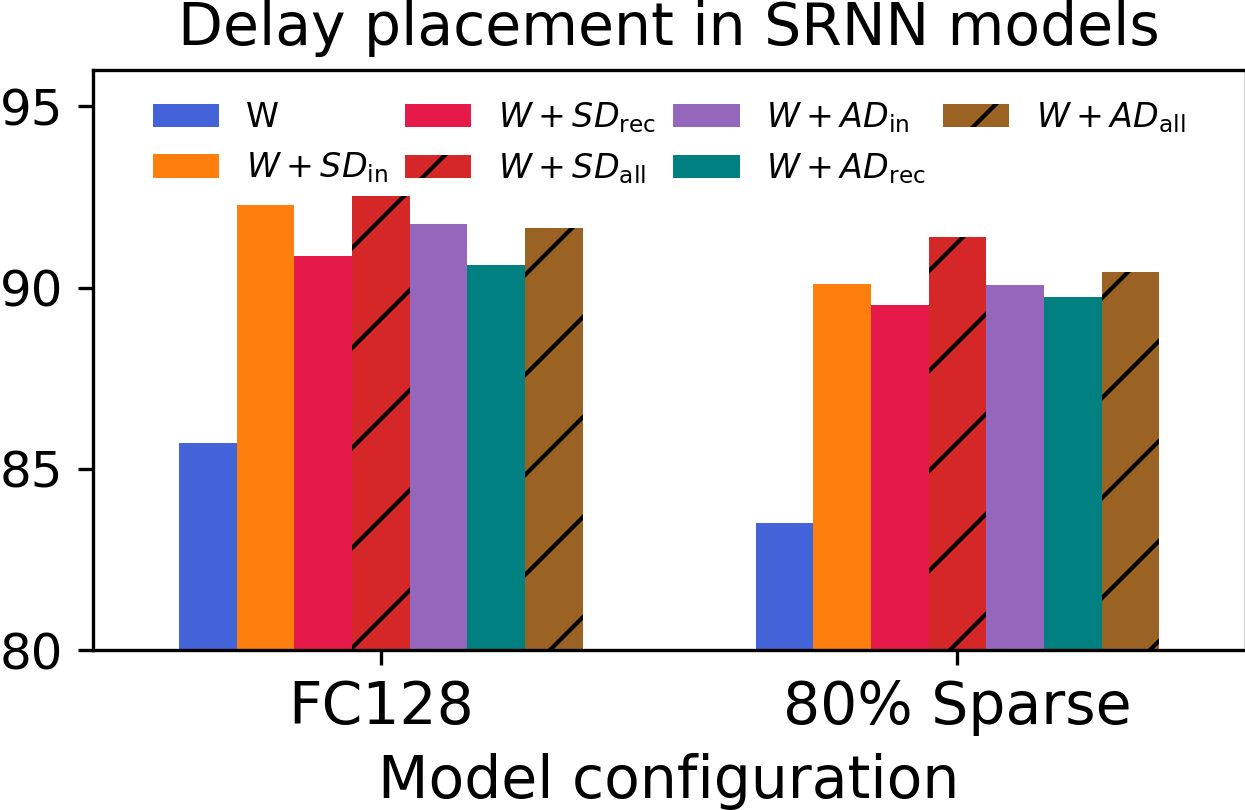}%
		\label{figure:delay_placement_effect_in_srnn}}
	
	\caption{Control experiments illustrating the efficacy of delay-learning. (a, b) Sweep model sparsity for axonal (AD) and synaptic delays (SD), with fixed (W+F*) or learnable (W+L*) delay parameters. (c) Compares learning delays with fixed weights (F-W*) using the proposed method and \gls{bptt}. (d) Compares \glspl{srnn} with delays in the input, recurrent, and all connections.}
	\label{fig:comprehensive_delay_comparison}
\end{figure*}

To investigate, Figure \ref{figure:delay_placement_effect_in_srnn} presents ablation experiments examining the effect of input and recurrent delays on classification accuracy. Adding synaptic input delays or recurrent delays offers a 6.56\% and 5.2\% increase in accuracy respectively, while adding both offers 6.6\%. Axonal delays follow a similar pattern with 0.5\% lower accuracy on average. In sparse configurations, the inclusion of input and recurrent synaptic delays provides a 1.3\% improvement in classification accuracy compared to input delays alone. Under these conditions, the accuracy loss remains at 0.95\%, compared to a 3.4\% loss in \glspl{snn}, despite an 80\% reduction in parameters. Axonal delays exhibit a similar trend, albeit with smaller improvements. 

\begin{figure}[bt]
	\centering
	\includegraphics[width=\linewidth]{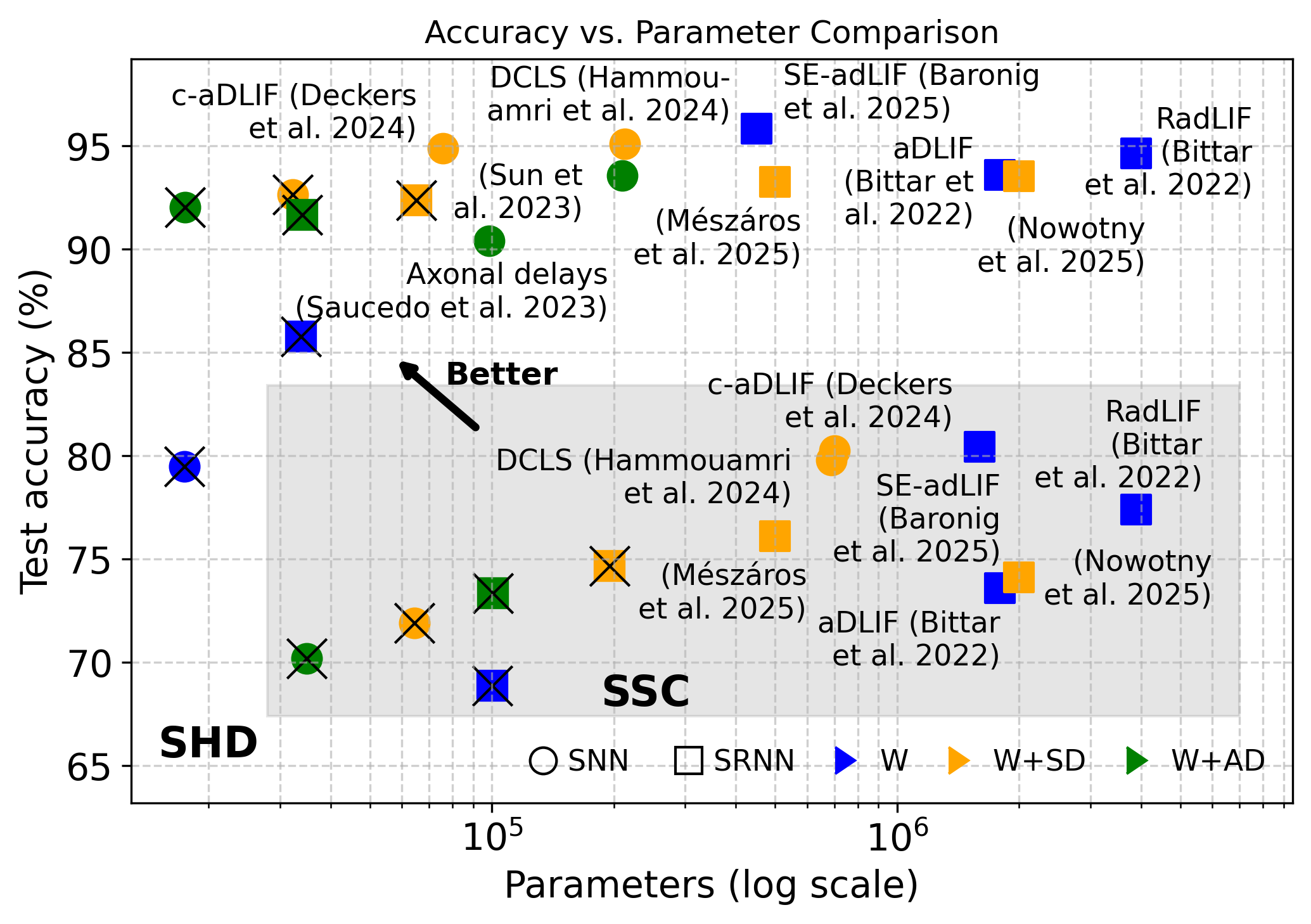}
	\caption{Comparison against offline methods on \gls{shd} and \gls{ssc} (gray background). Marker shape distinguishes model type, while color determines the type and presence of delays. Markers overlaid with an X correspond to the proposed online method}
	\label{figure:sota_comparison}
\end{figure}

The \gls{ssc} dataset poses a more challenging task for keyword spotting evidenced by a larger performance gap between models with fixed and learnable delays. In contrast, for the \gls{shd} dataset, fully connected models—whether feedforward or recurrent—exhibited negligible differences in accuracy. The classification accuracy of \glspl{snn} remains comparable to offline implementations. The versatility of \glspl{srnn} with synaptic and axonal delays leads up to 2.8\% and 3\% higher classification accuracy, respectively, compared to \glspl{snn}.

\section{Comparison and Discussion}
\label{section:discussion}
\Cref{figure:sota_comparison} compares our online method (marked with X) against competing state-of-the-art approaches on the \gls{shd} and \gls{ssc} datasets, where the latter are highlighted with a gray background. All competing methods are offline because they use \gls{bptt} or event-prop \cite{wunderlich_event-based_2021} to train their models, making them unsuitable for implementation on resource-constrained devices. Offline methods obtain highest accuracy with two \cite{hammouamri2024learning, baronig_advancing_2025, deckers_co-learning_2024} or even three hidden layers \cite{bittar_surrogate_2022}, while our method being approximate and constrained to a single layer achieves competitive performance within 3\% of the best offline results. Relative to the \gls{dcls} baseline, our approach achieves a 2.5\% lower test classification accuracy on \gls{shd} dataset. However, the loss of accuracy is offset by a 6.6$\times$ model size reduction, and since we only have one delay layer (three in the baseline), our model  also achieves 67\% lower inference latency.  Other competing solutions obtain marginally higher accuracy on the \gls{shd} albeit using a different model such as an adaptive \gls{srnn} \cite{baronig_advancing_2025, deckers_co-learning_2024, bittar_surrogate_2022} or event-based state space models \cite{schone_scalable_2024, soydan_s7_2024}. 

Similarly for the \gls{ssc} dataset, however, being more challenging we find the  more versatile \gls{srnn} offers 3\% higher accuracy, and therefore an advantage over feed-forward networks. However, having both more classes and varying recording conditions, our methods lags behind by 5.6\% in classification accuracy. We anticipate that incorporating multiple layers could further enhance accuracy on both datasets, but would be particularly beneficial for \gls{ssc}. 

\subsection{Efficacy of delay learning}
The major contribution to test classification accuracy is offered by the addition of fixed delays. As shown in Table \Cref{table:our-classification-accuracy}, co-learning weights and delays in dense situations offers only a modest 1-2\% accuracy improvement, as a result we conducted further experiments to further understand the imapct of delay learning. The efficacy of delay learning becomes more apparent under highly sparse conditions, where contributions up to 5-6\% over fixed delays can be observed in Figures \ref{fig:sweep_axonal} and \ref{fig:sweep_synaptic}.
The three-factor delay learning rules approximate backpropagation because the online recursive calculation of parameter updates, which enables their real-time implementation, relies entirely on historical information.  The comparison of the local learning rules to backpropagation in Figure \ref{fig:delay_learning} shows that the approximation holds for synaptic parameters with a gap of 2\%, while for axonal parameters it starts to break down as the gap widens to 8\%. The widened gap for axonal delays might be explained by the approximate parameter updates amassing into large errors, since each axonal parameter at the input neuron is derived from the accumulated contributions of all postsynaptic neurons. The significant drop in classification accuracy for F-W+AD in \Cref{fig:delay_learning} is likely due to the small number of trainable parameters, totaling 116. However, as learning under sparse conditions shows, when co-learned the impact of these errors could potentially have a regularization effect and overall yield acceptable results. These inaccuracies could potentially be mitigated by adopting alternative coding strategies, such as \gls{ttfs} coding, which would allow axonal delays to be assigned locally to the hidden neuron rather than the input neuron.

\subsection{Relevance for on-chip implementation}
Three factor delay learning rules offer a solution to the update locking problem \cite{pmlr-v70-jaderberg17a} enabling real-time on-device gradient calculations that closely approximate backpropagation. The resources of such systems are dominated by on-chip  \gls{sram} \cite{frenkel_reckon_2022, davies_loihi_2018} which accommodates both model parameters and the implementation of neural dynamics. Axonal delays scale linearly while synaptic delays scale quadratically with layer size which impacts the memory requirements for both parameter storage as well as their implementation. Beyond FIFO buffers for delaying input spikes, or ring buffers for delayed synaptic integration, the real-time calculation of delay parameter updates requires implementing the product between the eligibility trace and the surrogate kernel at the time the spike reaches the post-synaptic neuron. When no assumption can be made on the spike quantity for any given timestep the implementation relies on ring buffers. Their size is smallest when the input spike is delayed requiring a depth equal to the kernel length, but when pre-computing updates can reach a maximum equal to the larger between the kernel length and the maximum delay. Alternatively when activations are sparse, the post synaptic potentials can be buffered while keeping track of the spike times.
The added versatility can significantly reduce the size of the model while improving accuracy. Considering a typical quantisation of the setup in \Cref{section:experimental_setup}, with 8-bit weights, 5-bit delays, and 16-bit membrane potentials, the implementation of learnable axonal delays increases memory usage by 8.3\% of which 6.2\% are needed for learning. For synaptic delays this increases to 57.7\%, of which a third are needed for learning. Using a dense weights-only model as the baseline, a model that incorporates axonal or synaptic delays must reach at least 43\% and 74\% sparsity, respectively, in order to achieve the same memory footprint when both parameter storage and implementation overhead are considered.

\section{Conclusion}
In this work, we proposed three-factor learning rules for synaptic and axonal delays in \glspl{snn} and \glspl{srnn}, validated their implementation against state-of-the-art \gls{bptt}-based learning methods by demonstrating comparable performance, and studied the efficacy of delay learning through ablation experiments. 
Under online conditions, we demonstrated that delay learning improves classification accuracy by up to 20\% in \glspl{snn} and 12.5\% in \glspl{srnn} and is particularly advantageous for small networks, where co-learning weights and delays yield up to 14\% accuracy improvement over a weights-only model of equivalent size. While further research is required to scale to cognitively demanding tasks, online delay learning offers a practical solution to enhance embedded systems with limited on-chip \gls{sram} with adaptability without compromising accuracy, and enables trade-offs between implementation complexity and task performance. 

\section*{Author Contributions}
L.V. proposed the learning rule, associated experiments and their implementation in PyTorch, performed the literature review and wrote the initial version of the manuscript. N.T. provided supervision and funding. All authors discussed the results, reviewed, and edited the manuscript.

\section*{Funding}
This work was supported by the Hector Stiftung (grant no. 2304191) and the Field of Focus 2 at Heidelberg University (grant no. 33477).

\bibliographystyle{unsrt2authabbrvpp} 
\bibliography{references}

\end{document}